\documentclass[a4paper,11pt]{isoplex-report}
\ifdefined\pdfminorversion\pdfminorversion=7\fi
\usepackage{longtable,needspace,setspace,placeins}
\usepackage{etoolbox}
\newsavebox{\SagaTitleMetadata}
\makeatletter
\patchcmd{\maketitle}{%
  \begin{minipage}[b]{0.42\textwidth}
    \raggedleft\sffamily\fontsize{8}{11}\selectfont
    {\color{BrandNavy}\bfseries TECHNICAL REPORT}\par
    \vspace{3pt}
    {\color{TextGray}\@reportid\quad\textbar\quad Version \@reportversion\par\@date}
  \end{minipage}%
}{%
  \begin{lrbox}{\SagaTitleMetadata}\begin{minipage}[b]{0.42\textwidth}
    \raggedleft\sffamily\fontsize{8}{11}\selectfont
    {\color{BrandNavy}\bfseries TECHNICAL REPORT}\par
    \vspace{3pt}
    {\color{TextGray}\@reportid\quad\textbar\quad Version \@reportversion\par\@date}
  \end{minipage}\end{lrbox}\phantom{\usebox{\SagaTitleMetadata}}%
}{}{\PackageError{Saga}{Title metadata patch failed}{Check the locked template}}
\makeatother
\makeatletter
\AtBeginDocument{\def\LT@makecaption#1#2#3{%
  \LT@mcol\LT@cols{l}{\parbox[t]{\dimexpr\textwidth-2\tabcolsep\relax}{%
  \small\raggedright #1{\textbf{#2}. }#3\par\vspace{7pt}}}}}
\makeatother
\renewcommand{\arraystretch}{1.15}
\reportid{ISOPLEX-SAGA-TR-2026-001}
\reportversion{1.11.1}
\reportcompany{Isoplex Intelligence}
\reportaffiliation{\textsuperscript{1}Peking University\\
\textsuperscript{2}Peking Union Medical College\\
\textsuperscript{3}Isoplex Lab}
\reportshorttitle{Saga: Multimodal Patient World Model}
\reportkeywords{}
\reportnotice{}
\title{Saga: an in silico ontology for a multimodal patient world model}
\author{Youcheng Li\texorpdfstring{\textsuperscript{1,3}}{}, Fa Tian\texorpdfstring{\textsuperscript{3}}{}, Binbin Su\texorpdfstring{\textsuperscript{2,3}}{}}
\date{18 October 2026}
\begin{document}
\maketitle
\begin{abstract}
Simulating an individual's disease course requires connecting patient history to future events, clinical timing and disease morphology. Longitudinal prediction and multimodal representations exploit this information, but organizing future readouts around a common patient state and advancing trajectories without real intermediate observations remain central challenges. Here we introduce Saga, a multimodal patient world model that learns a shared state from patient histories to predict clinical events and generate multimodal images. Generated events and times are appended to the history and re-encoded to advance prediction. We studied 461,370 patients with 12 cancer types across 112 centres. Modality-specific paired evaluations linked future images to observed follow-up. With historical images held fixed, removing patient conditioning reduced event F1 from 0.716 to 0.591 and increased lesion-burden error from 11.8\% to 21.3\%. In free rollouts, eight-step event F1 was 0.672, compared with 0.553 for a longitudinal Transformer evaluated from the same origins, and timing error was lower by 8.7 days. In the internal test set, one-year progression AUROC was 0.882, with values of 0.871 and 0.856 in temporal and geographical external evaluations. These results show that shared patient states connect future phenotypes across observational scales with event-driven trajectory progression, sustaining comparative predictive advantages during rollout and discrimination in external clinical evaluation. Saga makes patient-level disease simulation a multimodal prediction task that can be tested against observed follow-up.
\end{abstract}

\par\medskip

\par\medskip
\section{Introduction}

Clinical assessment and disease research both require connecting a patient's current presentation to a disease course that has yet to unfold. Clinicians need to anticipate progression and identify changes worth monitoring; researchers seek to explain divergent trajectories and use that understanding to formulate hypotheses and design studies. Integrating observations, updating states and predicting changes at the patient level has therefore become an important direction in medical digital-twin research\citep{Laubenbacher2024DigitalTwins}. Cancer provides a longitudinal setting in which to test this capacity across scales: lung-cancer studies connect metastatic spread with continuing clonal evolution\citep{Hessey2026evolutionary}, while serial circulating-tumour-DNA measurements can identify impending relapse during radiological surveillance\citep{Abbosh2023tracking}. Clinical records, imaging and pathology capture different aspects of the same disease course but are acquired at different times and scales. Irregular sampling and missing information in clinical time series further complicate longitudinal modelling\citep{DeBrouwer2019}. A patient model must form an updatable representation from this fragmented history so that future events, clinical timing and disease phenotypes jointly describe a trajectory that can be tested against follow-up.

Longitudinal health-record models have established patient histories as a basis for multitask prediction and trajectory generation. Learned patient representations and end-to-end electronic health record models support prediction across clinical outcomes\citep{Miotto2016deep,Rajkomar2018scalable}, and pretrained record models further learn disease risks from previous clinical events\citep{Li2020,Rasmy2021}. Foresight uses clinical concepts to generate trajectories of future medical events\citep{Kraljevic2024foresight}, while ETHOS and Delphi model future health trajectories from longitudinal records\citep{Ethos2024,Shmatko2025}; DT-GPT further predicts longitudinal patient variables across clinical settings\citep{DTGPT2025}. These studies establish patient history as a basis for continuing future generation. Multimodal disease modelling further requires future events and observable phenotypes to be organized around a common patient representation that can update along a generated path.

Multimodal oncology models offer a complementary route for connecting clinical states with disease morphology. Integrating clinical, imaging, pathology and genomic information can improve predictions of prognosis or treatment response\citep{Chen2022,Jee2024,Vanguri2022}; ovarian-cancer studies further demonstrate the complementary contributions of radiological, histopathological and clinicogenomic features to risk stratification\citep{Boehm2022multimodal}. Vision--language models connect tissue morphology with clinical semantics\citep{Xiang2025,Titan2025}, and CONCH demonstrates classification, retrieval and captioning learned from paired pathology images and text\citep{Lu2024visual}. Conditional diffusion has also been used to generate medical training images and improve the out-of-distribution performance of classifiers\citep{Ktena2024}. The next question is whether predictive information in a patient's history can support prediction of both future events and disease morphology, making multimodal generation part of individual trajectory prediction.

World models describe changing environments through internal states, state transitions and generated observations, providing a framework for organizing these capabilities\citep{Ha2018,Hafner2025}. Medical world models have explored treatment-conditioned tumour evolution and longitudinal clinical-state prediction\citep{Yang2025,Ding2025CLARITY,Wang2026Chrono}; PULSE combines previous patient states with currently observed modalities for longitudinal multimodal alignment and generation\citep{Wu2026PULSE}. The recent HealthFlux preprint further organizes irregular multimodal observations into a health state that evolves between measurements and is revised when new observations arrive, supporting disease-risk simulation without subsequent observations\citep{Wang2026HealthFlux}. A multimodal patient world model connects two capabilities: reading future events and morphology from a common state, and advancing along its own generated history. Perturbing patient conditioning while holding historical images fixed tests the contribution of longitudinal state information to prediction of observed follow-up phenotypes. Continuing prediction after appending generated events and their times to the history measures capability and error propagation without real intermediate records. Together, these tests define an evaluable standard for patient models: predictions should depend on information about the individual and retain value as the trajectory advances without stepwise support from observed intermediate records.

Here we introduce Saga, a patient world model that learns a shared predictive representation of multimodal patient histories (\emph{in silico ontology}; Fig.~\ref{fig:framework}). Conditioned on treatment and time, the model predicts a future state and produces clinical events and multimodal images through clinical readouts and image-generation branches. Generated events and their times are appended to the history, which is re-encoded to advance prediction. Using a cohort of 461,370 patients with 12 cancer types from 112 centres, we evaluate future phenotypes, free-running trajectories and external clinical prediction.

\section{Methods}
\Needspace{6\baselineskip}\subsection{Cohort, partitions and available history}\label{meth:cohort}\label{meth:eligibility}\label{meth:history}

The cohort is organized into development, internal test, temporal held-out and geographically external sets. Cohort partitions and data coverage are presented in Table~\ref{tab:baseline}, cancer composition in Table~\ref{tab:cancerdata}, and centre distributions and the longitudinal prediction design in Fig.~\ref{fig:cohort}. Clinical-task populations are shown in Table~\ref{tab:tasks}, and longitudinal imaging pairs in Fig.~\ref{fig:imaging}a.

Patient counts are summarized by centre and cancer type. Coverage for a data category is the number of patients with that category divided by the corresponding patient population; categories may overlap. Core-centre summaries combine development, internal test and temporal external sets, with geographical external centres summarized separately. Clinical record density is the number of clinical event records divided by the patient count; within-centre cancer composition uses the centre patient count as its denominator. These descriptive patient- or centre-level summaries are reported separately from task-specific evaluation populations.

{
Let $E_i$, $t_i$ and $v_i$ denote event content, occurrence time and information-availability time, respectively. Available history at prediction origin $t_j$ is defined as
\begin{equation}
H_j=\{(E_i,t_i,v_i):t_i\leq t_j,\ v_i\leq t_j\}.
\end{equation}
This definition restricts inputs to information available at prediction time. A result obtained after the origin is excluded from that history even when the underlying event occurred earlier\citep{Kapoor2023Leakage}.
}

Patients are included if they have documented evidence of a tumour diagnosis, linkable patient identifiers and chronologically orderable clinical records. Clinical, imaging and pathology records are linked by patient, with duplicate records and cross-centre identities consolidated. Records that cannot be linked reliably, have conflicting key dates or lack the reference outcomes required for a task are excluded from that evaluation. Eligibility sets are established separately for clinical prediction, longitudinal imaging and trajectory evaluation; availability of all four imaging modalities is not required for inclusion in the overall cohort.\citep{Benchimol2015RECORD,Collins2024TRIPOD}

Patients are assigned to development and evaluation sets before prediction origins and observation pairs are generated, with all records from each patient retained in the same partition. Geographically external evaluation holds out entire centres. Temporal holdout includes subsequently eligible patients from development centres after a prespecified date cutoff, excluding patients already assigned to development or internal testing. Development uses only training information available before the cutoff. Each task defines origins at eligible encounters and truncates history by information-availability time; treatment inputs include only treatments and plans documented at the origin.

Diagnoses, treatments and laboratory measurements are mapped to common vocabularies and units; continuous-variable standardization parameters are estimated from the training set. Missingness indicators and observation masks represent missing items, and missing modalities are not interpreted as normal findings. Examinations and specimen collections are linked to clinical events by date, and patches retain their specimen and patient identifiers.

\Needspace{6\baselineskip}\subsection{Tasks, clinical endpoints and evaluation metrics}\label{meth:labels}\label{meth:eventmetrics}\label{meth:survivalmetrics}

The main clinical tasks are next event, 90-day events, one-year progression and overall survival (OS), reported using NDCG@5, AUPRC, AUROC and C-index, respectively (Fig.~\ref{fig:clinical}a). NDCG@5 evaluates the top-five event ranking\citep{Jarvelin2002NDCG}, AUPRC and AUROC evaluate classification discrimination\citep{Saito2015PR}, and the C-index evaluates survival ranking\citep{SurvivalMetrics}. Analyses across horizons retain AUROC for the next major clinical event and AUPRC for progression or recurrence, evaluated at 30, 90 and 180 days and 1, 2, 3 and 5 years (Fig.~\ref{fig:external}). Other tasks comprise PFS, DFS/RFS, first response, next treatment and treatment-line switching, using the respective classification, ranking or timing-error metrics (Table~\ref{tab:tasks}).

Progression uses progression recorded during clinical follow-up as its endpoint, whereas OS uses all-cause death\citep{FDA2018Endpoints}. Progression and progression-or-recurrence tasks use their respective clinical labels.

PFS uses progression or death as its endpoint, and DFS/RFS uses recurrence or death. First response uses the category recorded at a post-treatment response assessment. Next treatment uses the subsequently initiated regimen or treatment category, and treatment-line switching evaluates time to the start of the next treatment line.

For the additional metrics in Table~\ref{tab:tasks}, next-treatment Recall@5 measures coverage of reference treatment labels by the five highest-ranked candidates. First-response classes are complete response (CR), partial response (PR), stable disease (SD) and progressive disease (PD); macro-F1 is the equally weighted mean of class-specific F1 scores. Median absolute error for treatment-line switching is the median of the absolute differences between individual predicted and reference times.

Cancer-specific survival evaluation also reports time-dependent AUC and the integrated Brier score (IBS), describing discrimination at a specified time and prediction error integrated over time, respectively (Fig.~\ref{fig:external}d)\citep{Heagerty2000TimeROC,Graf1999Brier}.

\Needspace{6\baselineskip}\subsection{Shared patient state, transitions and clinical readouts}\label{meth:state}\label{meth:transition}\label{meth:readouts}

Saga organizes diagnoses, treatments, imaging, pathology, laboratory measurements and clinical text into event tokens. Content, type/modality, age and time-gap embeddings form the longitudinal input, which a Transformer with grouped temporal causal attention encodes into the shared patient state $S_j=q_\phi(H_j)$. The transition is conditioned on the current state, treatment $A_j$ and clinical interval $\delta_j$:
\begin{equation}
\begin{aligned}
\widehat{\Delta t}_j&=T_\rho(S_j,A_j),\\
\delta_j&=
\begin{cases}
\Delta t^{\mathrm{specified}},&\text{fixed-horizon prediction},\\
\widehat{\Delta t}_j,&\text{next-event prediction},
\end{cases}\\
\widehat S_{j+1}^{(q)}&\sim p_\theta(S\mid S_j,A_j,\delta_j).
\end{aligned}
\end{equation}
$S_j$ is a learned predictive representation, and $q$ indexes samples. For next-event prediction, a timing head first reads the current state and treatment to produce the waiting time used in the transition; fixed-horizon tasks instead use the specified interval. Clinical heads read the current or predicted state, while modality mappings read the predicted state to generate future images, supporting multiple tasks from the same patient history\citep{Ha2018,Hafner2025}.

\begin{figure}[!htbp]
\centering
\makebox[\textwidth][c]{%
\begin{minipage}{174mm}
\refstepcounter{figure}\label{fig:framework}
\noindent{\large\bfseries Fig. \arabic{figure}\quad A multimodal patient world model connects longitudinal history, future readouts and event feedback}\par
\begin{center}\includegraphics[width=\linewidth,height=0.73\textheight,keepaspectratio]{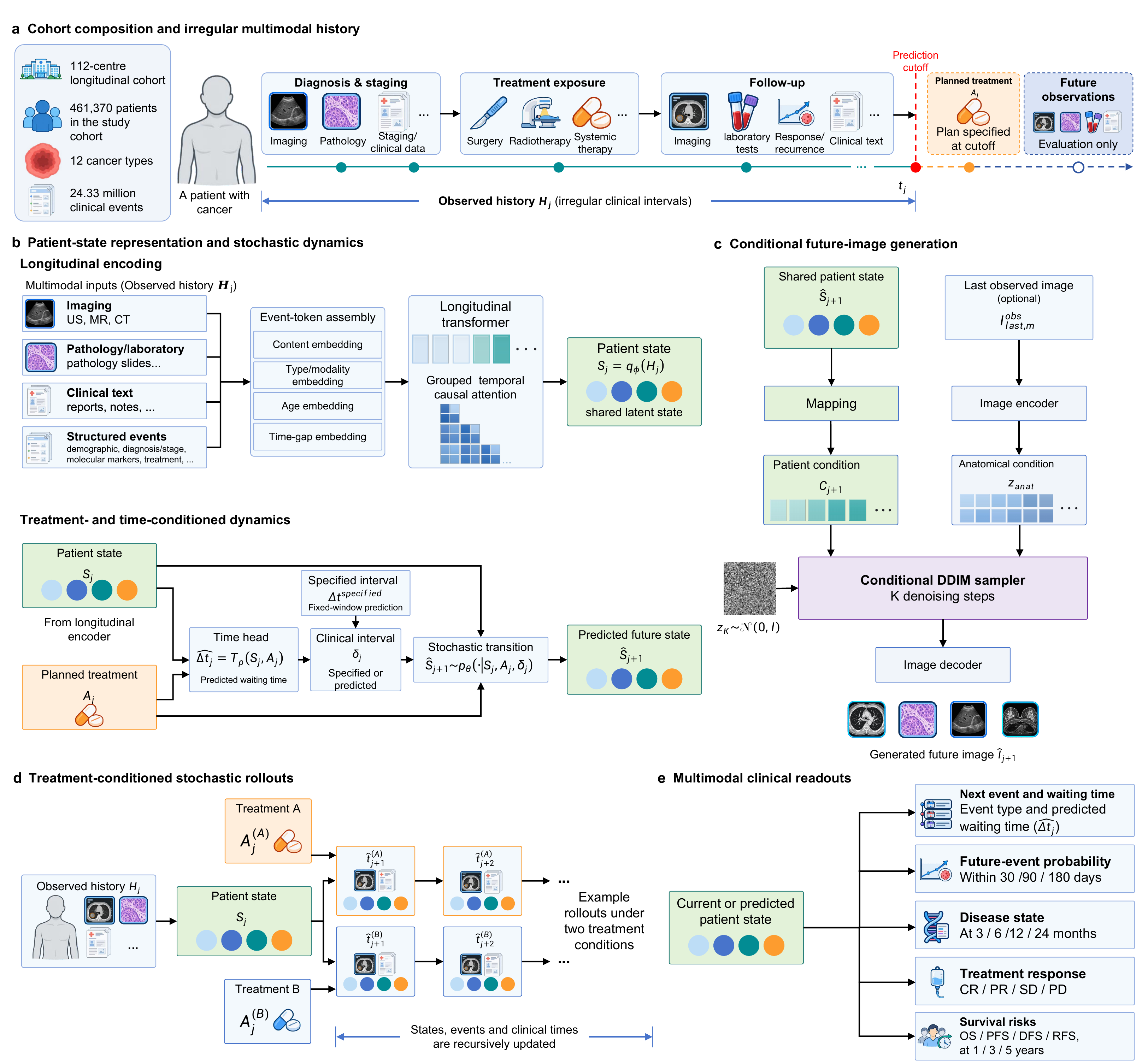}\end{center}
{\fontsize{8.5}{10.4}\selectfont a, Diagnoses, imaging, pathology, treatments, laboratory measurements and text records form a longitudinal patient history. Observations after the prediction origin provide supervision or evaluation. b, History is encoded into a shared patient state; future states are sampled from transitions conditioned on treatment known at the origin and the clinical interval. c, The predicted patient state is mapped to patient conditioning, while available historical images are encoded as anatomical conditioning. These two inputs guide DDIM sampling and decoding to generate ultrasound, MRI, CT and pathology images. d, Generated events and their times are appended to history for re-encoding; images remain downstream readouts. e, Evaluation interfaces for events, disease states, treatment responses and survival. Colours distinguish architectural modules.\par}
\end{minipage}}
\end{figure}

A visual encoder extracts imaging features, which a connector projects into the clinical-token embedding space for joint encoding with text and structured information within the same time block\citep{Liu2024LLaVA}. Longitudinal encoding uses the grouped temporal causal attention in Fig.~\ref{fig:framework}b, preventing earlier origins from accessing later time blocks.

\FloatBarrier

\Needspace{6\baselineskip}\subsection{Modality conditioning mappings and image generation}\label{meth:mapping}\label{meth:diffusion}

Image generation has separate patient-conditioning and historical-image pathways. Modality mappings for ultrasound, MRI, CT and pathology patches transform the same predicted state into patient conditions $C_{j+1}^{(m)}$; available historical images pass through image encoders to produce anatomical conditions $\mathcal C_j^{(m)}$:
\begin{equation}
\begin{aligned}
C_{j+1}^{(m)}&=g_{\omega_m}(\widehat S_{j+1}),\\
I_{j+1}^{(m)}&\sim p_\psi^{(m)}(I\mid C_{j+1}^{(m)},\mathcal C_j^{(m)}),
\quad m\in\{\mathrm{US},\mathrm{MRI},\mathrm{CT},\mathrm{path}\}.
\end{aligned}
\end{equation}
Both conditioning inputs guide denoising, and each branch generates images from initial noise through conditional DDIM sampling and decoding\citep{Song2021,Rombach2022}. Historical-image conditioning is optional, and pathology outputs are local patches. All four modalities follow the same conditional-generation procedure while retaining modality-specific encoding and decoding branches; quantitative imaging evaluation in this report covers ultrasound, MRI and pathology.

Modality mappings transform the predicted patient state into patient-condition tokens, while projected historical-image features form image-condition tokens. The two conditions enter the denoising network through cross-attention, representing patient state and historical-image content separately\citep{Rombach2022,Ye2023IPAdapter}.

\Needspace{6\baselineskip}\subsection{Training and sampling}\label{meth:training}

Clinical training combines event-classification, timing-prediction and survival objectives. An image autoencoder encodes observed follow-up images into latent variables; conditioned on patient state and historical imaging, the denoising network predicts Gaussian noise added to these variables using a mean-squared-error loss\citep{Rombach2022}. Records missing a label type or target image contribute only to objectives for which targets are available.

Checkpoint and hyperparameter selection are restricted to development data.

\Needspace{6\baselineskip}\subsection{Longitudinal imaging pairs and generation evaluation}\label{meth:preprocessing}\label{meth:imageoutcomes}

Longitudinal imaging evaluation uses observed follow-up from the same patient and modality as reference at 6, 12 and 24 months. Counts refer to longitudinal observation pairs, with pathology counted as patch pairs (Fig.~\ref{fig:imaging}a). Ultrasound uses lesion-change MAE, MRI uses burden-change MAE, and pathology uses representation cosine similarity and phenotype macro-F1 (Fig.~\ref{fig:imaging}b--d). Cosine similarity, $\operatorname{cos}(u,v)=u^\top v/(\|u\|\|v\|)$, describes alignment of representation vectors; macro-F1 is the equally weighted mean of category-specific F1 scores. These metrics describe continuous changes, tissue representations and phenotype classification, respectively.

Aggregate imaging evaluation also reports SSIM, PSNR and LPIPS for structural, intensity and perceptual similarity\citep{Wang2004SSIM,Wang2009MSE,Zhang2018LPIPS}, together with major-lesion-error and image-evaluability proportions (Fig.~\ref{fig:imaging}e).

Images are paired within the same patient, modality and anatomical site; generated and reference images use consistent evaluation regions and preprocessing.

MAE is the mean absolute difference between predicted and reference measurements, $N^{-1}\sum_i|\widehat y_i-y_i|$, with the same unit as the measurement. Fig.~\ref{fig:imaging}b presents ultrasound lesion-change and MRI burden-change errors separately, and Fig.~\ref{fig:imaging}f presents lesion-burden error in the conditioning experiment; all are expressed as percentages. Pathology evaluation uses consistent representation extraction for generated and observed images.

PSNR is defined by the image dynamic range $L$ and mean squared error as $\mathrm{PSNR}=10\log_{10}(L^2/\mathrm{MSE})$\citep{ImageMetrics}; LPIPS measures perceptual differences using pretrained network features and learned distance weights\citep{LPIPSCode}. SSIM and PSNR are evaluated in spatially corresponding image regions.

\Needspace{6\baselineskip}\subsection{Event feedback and free multistep rollout}\label{meth:rollout}

Free rollout starts from observed history and updates that history with generated events and occurrence times. Let $\widetilde H^{(0)}=H_j$, $S^{(0)}=q_\phi(H_j)$ and $\widehat t^{(0)}=t_j$. The recursion is
\begin{equation}
\begin{aligned}
\widehat S^{(r+1)}&\sim p_\theta(S\mid S^{(r)},A^{(r)},\delta^{(r)}),\\
\widehat t^{(r+1)}&=\widehat t^{(r)}+\delta^{(r)},\\
\widetilde H^{(r+1)}&=\operatorname{Append}(\widetilde H^{(r)},\widehat E^{(r+1)},\widehat t^{(r+1)}),\\
S^{(r+1)}&=q_\phi(\widetilde H^{(r+1)}).
\end{aligned}
\end{equation}
The predicted state supports current clinical and imaging readouts; the state re-encoded after event write-back supports the next transition. Observed historical images remain fixed at the origin, and generated images are downstream readouts that neither enter history nor update the clinical state. Step count denotes recursion depth, whereas elapsed clinical time is determined by generated intervals.

Figure~\ref{fig:rollout}a,b presents results at 1, 2, 4 and 8 evaluation depths. Event F1 compares Saga, the longitudinal Transformer, direct multihorizon prediction and a frozen-latent-state model; timing MAE compares the first two models. Event F1 is \mbox{$2\mathrm{TP}/(2\mathrm{TP}+\mathrm{FP}+\mathrm{FN})$}, where TP, FP and FN denote matched events, unmatched predicted events and unmatched reference events, respectively. Timing MAE is the mean absolute difference between predicted and reference occurrence times, expressed in days. Normalized edit distance, dynamic time warping (DTW) distance and patient-feature drift describe trajectory structure by measuring differences between event sequences, aligned trajectories and patient features, respectively (Fig.~\ref{fig:rollout}c). For recursive models, depth denotes the number of rollout steps.

Model comparisons use the same origin histories and reference follow-up; rollout inputs are limited to information available at the origin and events and times generated by the model itself.

\Needspace{6\baselineskip}\subsection{Comparators, ablations and conditional responses}\label{meth:baselines}

Clinical model comparisons include Saga, a clinical baseline, text-only and image-only models, a static multimodal model and a longitudinal Transformer (Fig.~\ref{fig:clinical}a). State-model comparisons include non-recurrent direct prediction, a frozen latent state, and removal of longitudinal history, treatment/action tokens or explicit time encoding (Fig.~\ref{fig:clinical}b). Component ablations compare linear mapping, no mapping, and removal of imaging, text, laboratory measurements, time encoding or pretraining (Fig.~\ref{fig:clinical}c), with each group referenced to its corresponding full model. Direct multihorizon prediction in Fig.~\ref{fig:rollout} is a separate comparator. Removal or exchange of patient conditions and changes to treatment conditions evaluate conditional responses in event and imaging readouts (Fig.~\ref{fig:imaging}f). State-model and component comparisons evaluate event F1 and timing MAE under each control condition; Fig.~\ref{fig:rollout} evaluates rollout performance at different depths.

Text-only and image-only models restrict the input modality. The static multimodal model aggregates and fuses features available at the origin without modelling event order, whereas the longitudinal Transformer encodes history over time and produces task predictions.

Non-recurrent direct prediction removes subsequent state recursion, whereas the frozen-latent-state comparator retains a fixed latent representation.

Progression-risk shifts describe changes in model readouts under altered conditioning and do not estimate causal treatment effects.

\Needspace{6\baselineskip}\subsection{Preliminary clinician assessment}\label{meth:readers}

Two physicians, each with more than ten years of clinical experience and experience in integrated interpretation of multimodal clinical information, performed an initial assessment of observed follow-up, Saga-generated trajectories and contradictory controls. They evaluated plausibility, image--text--time consistency, major contradictions and uncertainty or unevaluability, with inter-reader agreement summarized by $\kappa$ (Fig.~\ref{fig:rollout}d,e). These dimensions describe clinical plausibility, consistency across readouts and evaluability.

\Needspace{6\baselineskip}\subsection{External evaluation and population strata}\label{meth:external}

Hospital evaluation includes internal testing, temporal holdout, geographical holdout and leave-one-centre-out evaluation (Fig.~\ref{fig:external}a). Leave-one-centre-out results are summarized by their mean and between-centre standard deviation. Cancer-specific routine and external evaluations are shown in Fig.~\ref{fig:external}b,c, with additional cancer-specific metrics in Fig.~\ref{fig:external}d. Age, sex and stage results are presented in Table~\ref{tab:subgroups}.

\Needspace{6\baselineskip}\subsection{Statistical summaries}\label{meth:statistics}

This report describes results through point estimates and within-experiment model differences, with standard deviations describing between-centre variation. Table~\ref{tab:tasks} also reports 95\% confidence intervals for secondary tasks. Descriptive comparisons use the direction and magnitude of differences.

\FloatBarrier
\Needspace{10\baselineskip}
\section{Results}

\Needspace{6\baselineskip}\subsection{Data foundation and patient-state modelling}

\begin{figure}[!htbp]
\centering
\makebox[\textwidth][c]{%
\begin{minipage}{174mm}
\refstepcounter{figure}\label{fig:cohort}
\noindent{\large\bfseries Fig. \arabic{figure}\quad Data distributions and longitudinal prediction study design}\par
\begin{center}\includegraphics[width=\linewidth,height=0.73\textheight,keepaspectratio]{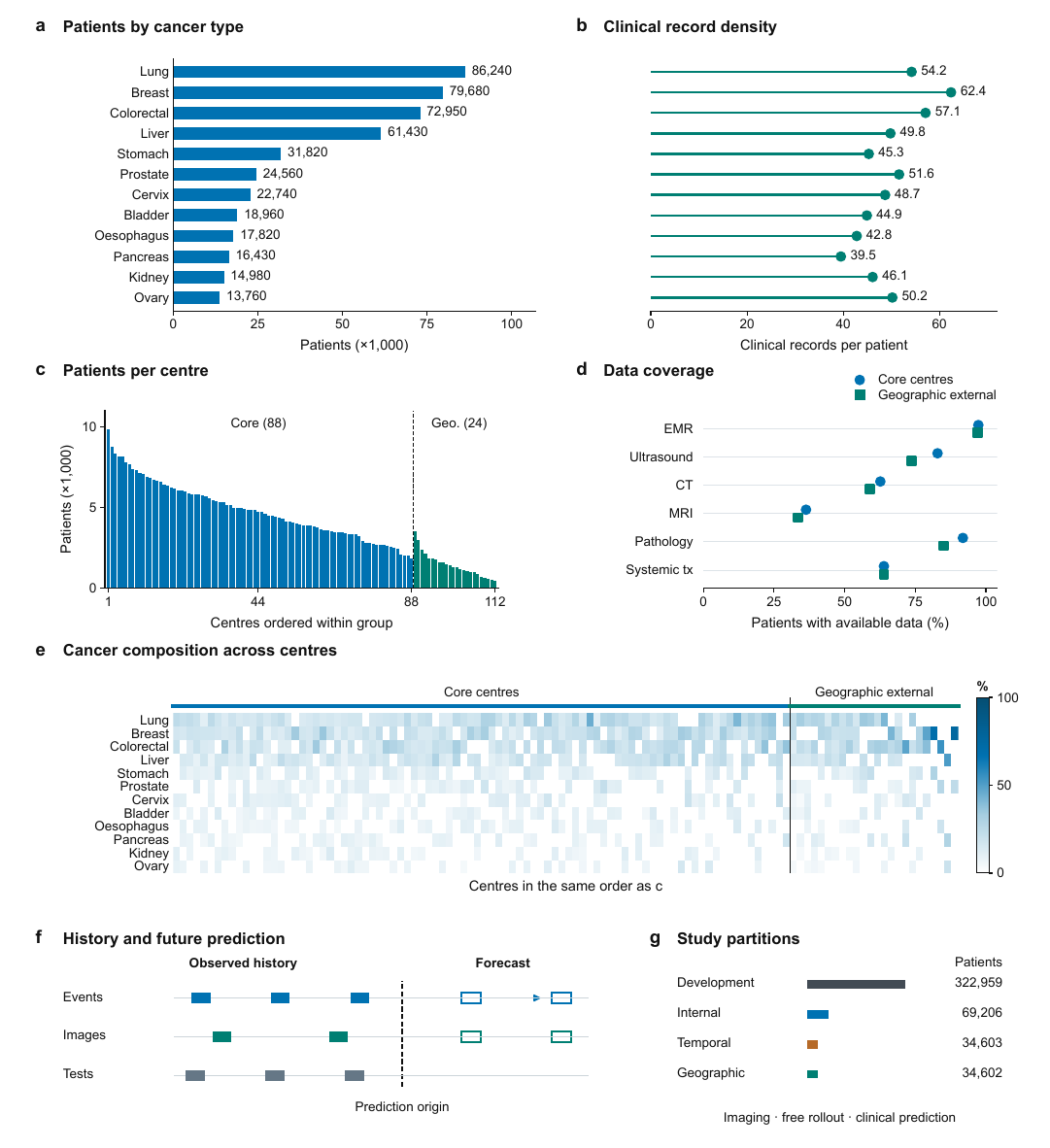}\end{center}
{\fontsize{8.5}{10.4}\selectfont a, Patient counts by cancer type. b, Mean clinical record density for each cancer, calculated as clinical event records divided by patients; cancer order matches a. c, Patients per centre, sorted in descending order separately within core and geographic external centres. Each bar represents one centre. d, Availability of electronic medical records, ultrasound, CT, MRI, pathology and systemic-treatment records in the pooled core centres and geographic external set. Each percentage uses the patient count of the corresponding group as its denominator; data categories may overlap. e, Cancer composition within centres. Columns follow the centre order in c; colour denotes the proportion of each centre's patients belonging to that cancer type. f, Conceptual multimodal history available before the prediction origin and future event, time and image readouts. Filled and open symbols denote observed history and future readouts, respectively. g, Patient counts in the development, internal test, temporal external and geographic external sets, and the imaging prediction, free rollout and clinical prediction tasks. Centre sizes and cancer composition are centre-level aggregates; coverage counts patients with available data. The core-centre pool comprises the development, internal test and temporal external sets. Data availability is distinct from evaluable longitudinal imaging pairs.\par}
\end{minipage}}
\end{figure}

Saga organizes patient histories into an updatable shared state from which future clinical events and disease morphology at different scales can be read out (Fig.~\ref{fig:framework}). The study includes 461,370 patients with 12 cancer types across 112 medical centres. Development, internal held-out, temporal external and geographical external sets contain 322,959, 69,206, 34,603 and 34,602 patients, respectively (Table~\ref{tab:baseline}; Fig.~\ref{fig:cohort}g). The first three partitions comprise 426,768 patients from 88 core centres; the geographical external set comprises a further 24 centres.

The cohort contains 24,331,394 clinical event records and 130,010 death events. Lung, breast, colorectal and liver cancers contribute the largest patient groups; cancer-specific patient counts, centre coverage and data availability are shown in Table~\ref{tab:cancerdata}. Clinical record density by cancer and within-centre cancer composition are shown in Fig.~\ref{fig:cohort}b,e. Patient counts per centre range from 1,831 to 9,878 in core centres and from 478 to 3,521 in geographical external centres, illustrating variation in centre size (Fig.~\ref{fig:cohort}c).

Electronic medical records cover 448,930 patients (97.3\%), and clinical imaging data (excluding pathology) are available for 412,476 patients (89.4\%). Patient coverage is 82.2\%, 62.4\%, 36.1\% and 91.3\% for ultrasound, CT, MRI and pathology, respectively. Systemic-treatment data cover 294,823 patients (63.9\%). Coverage in core and geographical external centres is shown in Table~\ref{tab:coverage} and Fig.~\ref{fig:cohort}d. Data categories may overlap within patients; Fig.~\ref{fig:cohort}f illustrates the longitudinal prediction paradigm using multimodal histories. Generative branches cover ultrasound, MRI, CT and pathology; quantitative imaging evaluation in this report covers ultrasound, MRI and pathology patches.

\begin{table}[!htbp]
\centering
\begin{minipage}{\textwidth}
{\fontsize{9}{11.5}\selectfont\setlength{\tabcolsep}{2pt}\renewcommand{\arraystretch}{1.25}
\fontsize{8.5}{11.0}\selectfont

\caption{Cohort partitions and data coverage.}\label{tab:baseline}\label{tab:coverage}
\begin{tabular}{@{}>{\raggedright\arraybackslash}p{0.26\linewidth}>{\raggedright\arraybackslash}p{0.22\linewidth}>{\raggedright\arraybackslash}p{0.22\linewidth}>{\raggedright\arraybackslash}p{0.22\linewidth}@{}}
\toprule
\textbf{Cohort} & \textbf{Patients} & \textbf{Centres} & \textbf{Role}\\
\midrule

\multicolumn{4}{@{}l}{\textbf{A. Cohort partitions}}\\
Development & 322,959 & 88 & Training and tuning\\
Internal test & 69,206 & 88 & Locked test\\
Temporal external & 34,603 & 88 & Future-time evaluation\\
Geographical external & 34,602 & 24 & External-centre evaluation\\
\midrule\multicolumn{4}{@{}l}{\textbf{B. Data coverage}}\\
\textbf{Measure} & \textbf{Overall} & \textbf{Core centres combined} & \textbf{Geographical external centres}\\
Patients & 461,370 & 426,768 & 34,602\\
Centres & 112 & 88 & 24\\
Clinical event records & 24,331,394 & 22,514,707 & 1,816,687\\
Death events & 130,010 & 120,206 & 9,804\\
Electronic medical records & 448,930 (97.3\%) & 415,335 (97.3\%) & 33,595 (97.1\%)\\
Ultrasound & 379,220 (82.2\%) & 353,696 (82.9\%) & 25,524 (73.8\%)\\
CT & 287,732 (62.4\%) & 267,344 (62.6\%) & 20,388 (58.9\%)\\
MRI & 166,707 (36.1\%) & 155,107 (36.3\%) & 11,600 (33.5\%)\\
Pathology & 421,411 (91.3\%) & 392,000 (91.9\%) & 29,411 (85.0\%)\\
Systemic treatment & 294,823 (63.9\%) & 272,718 (63.9\%) & 22,105 (63.9\%)\\

\bottomrule
\end{tabular}

\noindent Development, internal test and temporal external sets share the core centres; the geographical external set comprises separate centres. In B, data categories are patients n (percentage of the column total) and may overlap. Coverage differs from counts of examinations, slides or patches. The core-centre column pools the first three sets. Clinical and death events are aggregate counts.\par}
\end{minipage}
\end{table}

\begin{table}[!htbp]
\centering
\begin{minipage}{\textwidth}
{\fontsize{9}{11.5}\selectfont\setlength{\tabcolsep}{2pt}\renewcommand{\arraystretch}{1.25}
\fontsize{8}{10.5}\selectfont

\caption{Cancer composition, clinical events and data availability.}\label{tab:cancerdata}
\begin{tabular}{@{}>{\raggedright\arraybackslash}p{0.112\linewidth}>{\raggedright\arraybackslash}p{0.065\linewidth}>{\raggedright\arraybackslash}p{0.12\linewidth}>{\raggedright\arraybackslash}p{0.148\linewidth}>{\raggedright\arraybackslash}p{0.125\linewidth}>{\raggedright\arraybackslash}p{0.125\linewidth}>{\raggedright\arraybackslash}p{0.125\linewidth}>{\raggedright\arraybackslash}p{0.105\linewidth}@{}}
\toprule
\textbf{Cancer} & \textbf{Centres} & \textbf{Patients} & \textbf{Clinical events} & \textbf{Patients with clinical imaging} & \textbf{Patients with pathology} & \textbf{Patients with systemic treatment} & \textbf{Death events}\\
\midrule

Lung & 96 & 86,240 & 4,674,208 & 81,066 & 78,478 & 59,506 & 33,634\\
Breast & 91 & 79,680 & 4,972,032 & 72,509 & 75,696 & 57,370 & 9,562\\
Colorectal & 86 & 72,950 & 4,165,445 & 62,737 & 67,114 & 48,876 & 16,778\\
Liver & 82 & 61,430 & 3,059,214 & 58,358 & 54,058 & 35,629 & 25,801\\
Stomach & 61 & 31,820 & 1,441,446 & 26,092 & 29,593 & 21,001 & 11,137\\
Prostate & 53 & 24,560 & 1,267,296 & 19,157 & 21,858 & 10,070 & 2,456\\
Cervix & 49 & 22,740 & 1,107,438 & 18,419 & 21,376 & 12,962 & 4,093\\
Bladder & 45 & 18,960 & 851,304 & 16,685 & 17,064 & 8,722 & 3,792\\
Esophagus & 47 & 17,820 & 762,696 & 15,503 & 16,394 & 11,405 & 7,306\\
Pancreas & 42 & 16,430 & 648,985 & 15,773 & 13,801 & 12,322 & 9,201\\
Kidney & 40 & 14,980 & 690,578 & 13,931 & 13,182 & 7,190 & 2,397\\
Ovary & 38 & 13,760 & 690,752 & 12,246 & 12,797 & 9,770 & 3,853\\
Overall & 112 & 461,370 & 24,331,394 & 412,476 & 421,411 & 294,823 & 130,010\\

\bottomrule
\end{tabular}

\noindent Cancer-specific centre counts include centres with at least one patient of that cancer type; a centre may contribute to several cancers. Clinical imaging includes ultrasound, CT, MRI and other clinical imaging data, excluding pathology. Patient coverage may overlap across clinical imaging, pathology and systemic-treatment data. Death events are aggregate counts over the observation period. Within-centre cancer composition is shown in Fig.~\ref{fig:cohort}e.\par}
\end{minipage}
\end{table}

\FloatBarrier
\Needspace{6\baselineskip}\subsection{Future imaging prediction and responses to patient conditioning}

\begin{figure}[!htbp]
\centering
\makebox[\textwidth][c]{%
\begin{minipage}{174mm}
\refstepcounter{figure}\label{fig:imaging}
\noindent{\large\bfseries Fig. \arabic{figure}\quad Longitudinal future-image prediction and responses to patient conditioning}\par
\begin{center}\includegraphics[width=\linewidth,height=0.73\textheight,keepaspectratio]{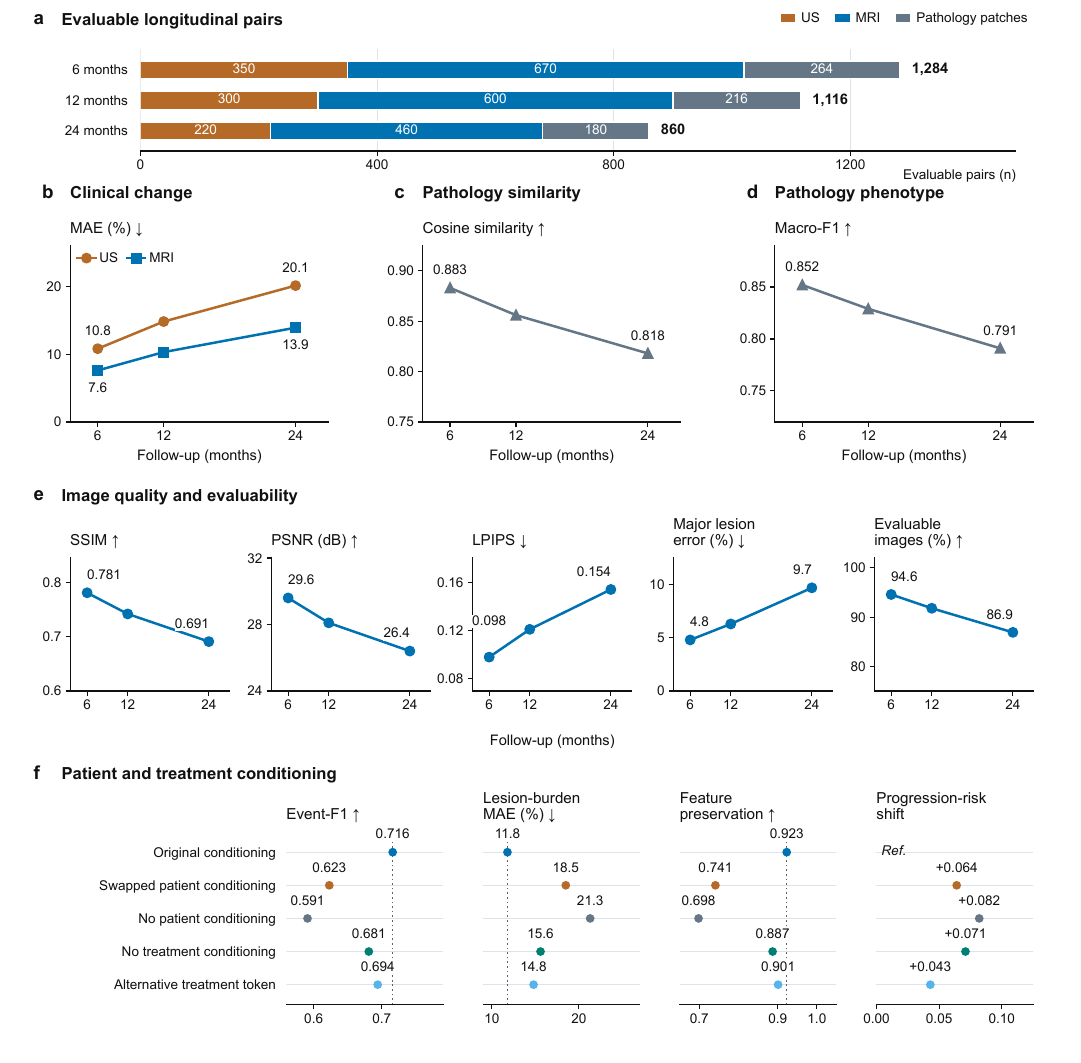}\end{center}
{\fontsize{8.5}{10.4}\selectfont a, Evaluable longitudinal pairs at 6, 12 and 24 months. Coloured segments in the stacked bars represent ultrasound, MRI and pathology-patch pairs; labels within segments show modality-specific counts, and labels at bar ends show totals. Counts refer to paired observations rather than unique patients. b, Mean absolute error (MAE) for ultrasound lesion change and MRI burden change. c, Representation cosine similarity for pathology patches. d, Pathology phenotype macro-F1. e, Structural similarity (SSIM), peak signal-to-noise ratio (PSNR), perceptual distance (LPIPS), major-lesion-error rate and image-evaluability rate in the aggregate imaging evaluation. f, Event-F1, lesion-burden MAE, feature preservation and shifts in progression-risk readouts under correct, swapped or removed patient conditioning, and removed or substituted treatment conditioning. Columns are aligned by conditioning setting. Dashed lines in the first three columns mark the estimates under correct conditioning; Ref. denotes the original reference in the risk-shift column. Points and curves show aggregate values. Panels b--e show 6-, 12- and 24-month follow-up windows, with the original scale retained for each metric; modality-specific and aggregate imaging metrics are presented separately. $\uparrow$ denotes higher is better; $\downarrow$ denotes lower is better. Risk shifts do not indicate the direction of benefit.\par}
\end{minipage}}
\end{figure}

Using paired follow-up observations from the same patients, we evaluated morphological change in ultrasound and MRI, and representation similarity and phenotype in pathology (Fig.~\ref{fig:imaging}b--d). The 6-, 12- and 24-month evaluations contain 1,284, 1,116 and 860 evaluable pairs, comprising 350, 300 and 220 ultrasound pairs, 670, 600 and 460 MRI pairs, and 264, 216 and 180 pathology-patch pairs (Fig.~\ref{fig:imaging}a). Pathology representation cosine similarity declined from 0.883 at 6 months to 0.818 at 24 months, and phenotype macro-F1 from 0.852 to 0.791 (Fig.~\ref{fig:imaging}c,d). In the evaluation of morphological change, ultrasound lesion-change MAE increased from 10.8\% to 20.1\%, and MRI burden-change MAE from 7.6\% to 13.9\% (Fig.~\ref{fig:imaging}b). Counts refer to paired longitudinal observations; pathology is counted at the patch-pair level.

Future-phenotype prediction deteriorated as the horizon extended. In the summary imaging evaluation, SSIM declined from 0.781 at 6 months to 0.691 at 24 months, LPIPS increased from 0.098 to 0.154, major lesion errors increased from 4.8\% to 9.7\%, and evaluability declined from 94.6\% to 86.9\% (Fig.~\ref{fig:imaging}e). Image fidelity, lesion errors and evaluability consistently indicate greater difficulty at longer horizons.

Conditioning comparisons with historical images held fixed test the contribution of patient information to future prediction. Under matched, swapped and removed patient conditions, event F1 was 0.716, 0.623 and 0.591, and lesion-burden MAE was 11.8\%, 18.5\% and 21.3\%, respectively (Fig.~\ref{fig:imaging}f). Removing patient conditioning reduced event F1 by 0.125 and increased burden error by 9.5 percentage points; swapping patient conditions produced corresponding changes of 0.093 and 6.7 percentage points. Removing treatment conditions also reduced event F1 to 0.681 and increased burden error to 15.6\%. With historical images held fixed, changes in patient identity and treatment conditioning affected prediction, supporting a contribution of patient information to event and morphology readouts. Shifts in progression-risk readouts and feature preservation further characterize responses to patient and treatment conditions (Fig.~\ref{fig:imaging}f).

\FloatBarrier
\Needspace{6\baselineskip}\subsection{Event feedback supports continuing trajectories with clinical coherence}

Free rollout appends generated events and times to the history and re-encodes it to update the patient state. Images are downstream readouts at the corresponding time points (Fig.~\ref{fig:framework}c,d).

\begin{figure}[!htbp]
\centering
\makebox[\textwidth][c]{%
\begin{minipage}{174mm}
\refstepcounter{figure}\label{fig:rollout}
\noindent{\large\bfseries Fig. \arabic{figure}\quad Predictive performance, trajectory structure and clinical assessment of free rollout}\par
\begin{center}\includegraphics[width=\linewidth,height=0.73\textheight,keepaspectratio]{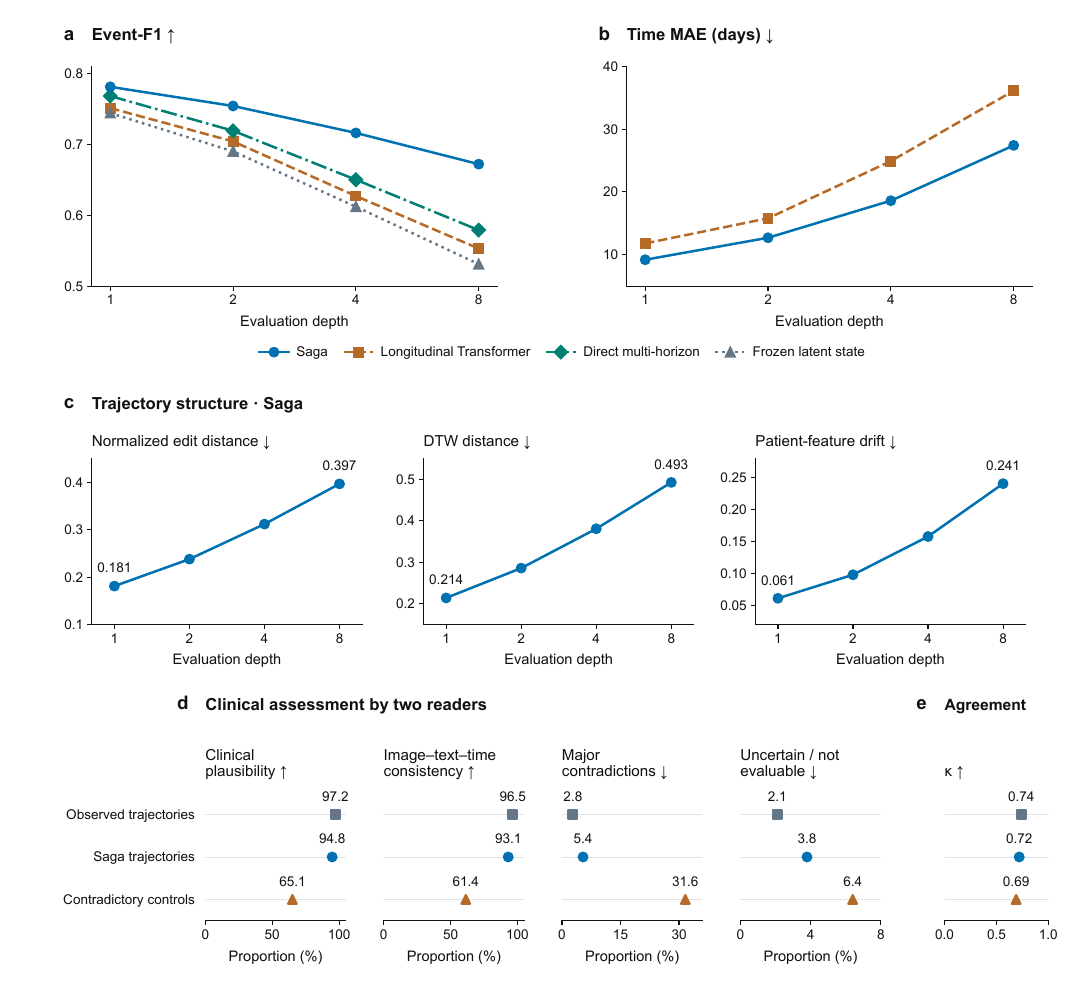}\end{center}
{\fontsize{8.5}{10.4}\selectfont a, Event-F1 for Saga, a longitudinal Transformer, direct multihorizon prediction and a frozen-latent-state model at 1-, 2-, 4- and 8-step evaluation depths. b, Event-time MAE in days for Saga and the longitudinal Transformer at the same evaluation depths. Panels a and b share the model legend; the horizontal axis denotes stepwise evaluation depth, corresponding to rollout steps for recursive models. c, Normalized edit distance, dynamic time warping (DTW) distance and patient-feature drift for Saga trajectories; each of the three columns retains its own metric scale. d, Clinical plausibility, image--text--time consistency, major contradictions, and uncertain or not-evaluable ratings by two physicians for observed trajectories, Saga trajectories and contradiction controls. The four columns are aligned across the same three trajectory groups, with each dimension assessed separately. e, Inter-reader agreement $\kappa$ for the three trajectory groups, aligned by row with panel d. Points show aggregate values. Lines in a--c connect adjacent evaluated step counts; d and e contain no error intervals. $\uparrow$ denotes higher is better; $\downarrow$ denotes lower is better. Higher $\kappa$ indicates greater inter-reader agreement.\par}
\end{minipage}}
\end{figure}

In the stepwise comparison, event F1 at 1, 2, 4 and 8 steps was 0.781, 0.754, 0.716 and 0.672 for Saga, compared with 0.751, 0.704, 0.627 and 0.553 for the longitudinal Transformer (Fig.~\ref{fig:rollout}a). The absolute F1 difference increased from 0.030 at one step to 0.119 at eight steps. At the eight-step evaluation depth, direct multihorizon prediction and frozen-latent-state controls achieved event F1 values of 0.579 and 0.531. Saga's timing MAE increased from 9.2 days at one step to 27.4 days at eight steps, remaining below the Transformer's corresponding errors of 11.8 and 36.1 days, with an eight-step reduction of 8.7 days (Fig.~\ref{fig:rollout}b).

Saga's trajectory-structure metrics describe how errors change with rollout depth. Between one and eight steps, normalized edit distance increased from 0.181 to 0.397, dynamic time warping distance from 0.214 to 0.493, and feature drift from 0.061 to 0.241. These results describe changes in Saga's sequence differences, trajectory differences and patient-feature deviations as rollout deepens (Fig.~\ref{fig:rollout}c).

A preliminary assessment by two physicians evaluated the clinical plausibility and cross-output consistency of recursive trajectories. Model trajectories achieved 94.8\% clinical plausibility, 93.1\% image--text--time consistency and a 5.4\% major-contradiction rate, compared with 97.2\%, 96.5\% and 2.8\% for real trajectories (Fig.~\ref{fig:rollout}d). Deliberately contradictory trajectories had a 31.6\% major-contradiction rate, providing a conflict reference. Inter-rater agreement for model trajectories was $\kappa=0.72$ (Fig.~\ref{fig:rollout}e), and the uncertain or unevaluable proportion was 3.8\%. The preliminary assessment complements quantitative evaluation with ratings of trajectory plausibility and cross-output coherence.

\FloatBarrier
\Needspace{6\baselineskip}\subsection{Multitask clinical prediction and patient-state component comparisons}

Clinical evaluation covers next-event prediction, 90-day events, one-year progression and overall survival (OS). Task-specific patient and event counts are presented in Table~\ref{tab:tasks}.

\begin{figure}[!htbp]
\centering
\makebox[\textwidth][c]{%
\begin{minipage}{174mm}
\refstepcounter{figure}\label{fig:clinical}
\noindent{\large\bfseries Fig. \arabic{figure}\quad Multitask clinical prediction and patient-state component comparisons}\par
\begin{center}\includegraphics[width=\linewidth,height=0.73\textheight,keepaspectratio]{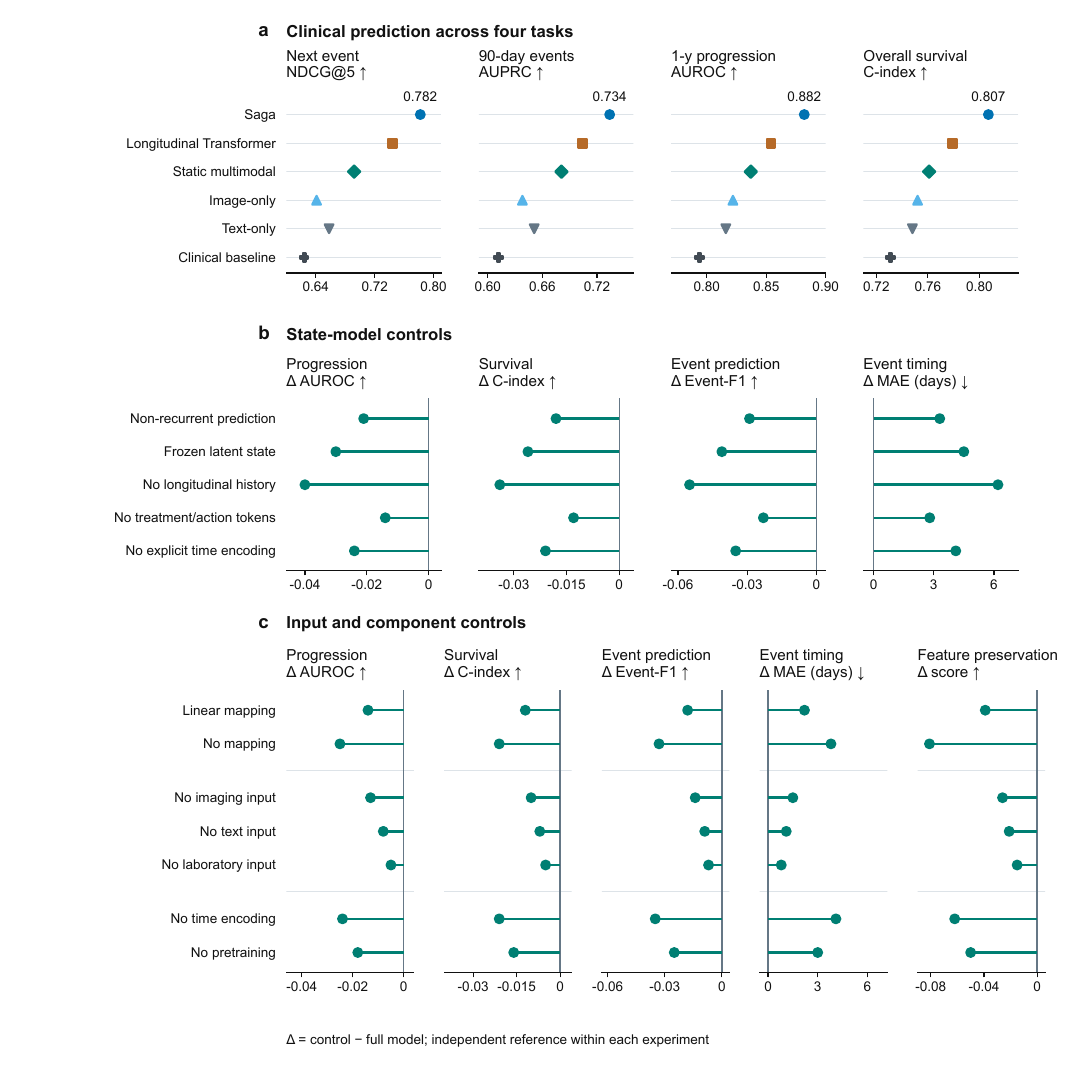}\end{center}
{\fontsize{8.5}{10.4}\selectfont a, Performance across six model classes on next-event, 90-day-event, one-year-progression and overall-survival tasks, measured by NDCG@5, AUPRC, AUROC and C-index, respectively. Each point shows an aggregate result for one model. b, Differences from the full model in progression AUROC, survival C-index, Event-F1 and event-time MAE for non-recursive direct prediction, a frozen latent state, and removal of longitudinal history, treatment/action tokens or explicit time encoding. c, Differences in five metrics for controls involving mapping, input modalities, time encoding and pretraining: progression AUROC, survival C-index, Event-F1, event-time MAE and feature preservation. Panels b and c use the full model from their respective experiments as the reference. Points show control minus full-model differences; thin segments connect each point to zero and do not represent confidence intervals. Event-time MAE differences are in days; other differences are dimensionless. $\uparrow$ denotes higher is better; $\downarrow$ denotes lower is better.\par}
\end{minipage}}
\end{figure}

Relative to the strongest listed comparator, the longitudinal Transformer, the full model increased next-event NDCG@5 from 0.744 to 0.782, 90-day event AUPRC from 0.704 to 0.734, one-year progression AUROC from 0.854 to 0.882, and OS C-index from 0.779 to 0.807 (Fig.~\ref{fig:clinical}a).

State-model comparisons describe clinical performance across prediction configurations. Relative to non-recurrent prediction, the full model improved one-year progression AUROC, event F1 and timing MAE by 0.021, 0.029 and 3.3 days, respectively; corresponding improvements over a frozen latent state were 0.030, 0.041 and 4.5 days (Fig.~\ref{fig:clinical}b). Removing longitudinal history reduced progression AUROC to 0.842 and event F1 to 0.661, while increasing timing MAE to 24.8 days. Removing explicit time encoding reduced progression AUROC to 0.858 and increased timing MAE to 22.7 days. Comparisons of mapping form, input modalities and pretraining are shown in Fig.~\ref{fig:clinical}c.

Secondary tasks included PFS prediction with a C-index of 0.791, first-response classification with a macro-AUROC of 0.864, next-treatment prediction with an NDCG@5 of 0.758, and treatment-line switching with a timing MAE of 21.3 days; DFS/RFS and other results are reported in Table~\ref{tab:tasks}.

\begin{table}[!htbp]
\centering
\begin{minipage}{\textwidth}
{\fontsize{9}{11.5}\selectfont\setlength{\tabcolsep}{3pt}\renewcommand{\arraystretch}{1.25}

\caption{Clinical-task evaluation populations and additional-task results.}\label{tab:tasks}
\begin{tabular}{@{}>{\raggedright\arraybackslash}p{0.24\linewidth}>{\raggedright\arraybackslash}p{0.16\linewidth}>{\raggedright\arraybackslash}p{0.15\linewidth}>{\raggedright\arraybackslash}p{0.18\linewidth}>{\raggedright\arraybackslash}p{0.21\linewidth}@{}}
\toprule

\multicolumn{5}{@{}l}{\textbf{A. Evaluation populations for primary tasks}}\\[3pt]
\textbf{Evaluation set} & \textbf{Next event, n} & \textbf{90-day event, n} & \textbf{1-year progression, n/events} & \textbf{OS, n/deaths}\\
\midrule
Internal test & 65,104 & 62,840 & 58,904 / 14,420 & 66,721 / 16,418\\
Temporal external & 32,730 & 31,076 & 28,820 / 7,234 & 33,650 / 9,441\\
Geographic external & 32,119 & 30,502 & 28,406 / 7,386 & 33,041 / 9,516\\
\midrule\multicolumn{5}{@{}l}{\textbf{B. Additional clinical tasks}}\\[3pt]
\textbf{Task} & \textbf{Primary metric} & \textbf{Estimate} & \textbf{95\% CI} & \textbf{Additional metric}\\
\midrule
Progression-free survival (PFS) & C-index & 0.791 & 0.786--0.796 & 1-year td-AUC 0.846\\
Disease-/recurrence-free survival (DFS/RFS) & C-index & 0.783 & 0.777--0.789 & 3-year td-AUC 0.852\\
Next-treatment prediction & NDCG@5 & 0.758 & 0.752--0.764 & Recall@5 0.846\\
First treatment response & Macro-AUROC & 0.864 & 0.858--0.870 & Macro-F1 0.721\\
Time to treatment-line switch & MAE (days) & 21.3 & 20.4--22.2 & Median absolute error 14.6 days\\

\bottomrule
\end{tabular}

\noindent In section A, progression counts are patients/events and OS counts are patients/deaths; task populations may overlap. Section B gives point estimates, confidence intervals and additional metrics for other tasks. First-response classes are CR, PR, SD and PD.\par}
\end{minipage}
\end{table}

\FloatBarrier
\Needspace{6\baselineskip}\subsection{Performance across hospitals, cancers and prediction horizons}

\begin{figure}[!htbp]
\centering
\makebox[\textwidth][c]{%
\begin{minipage}{174mm}
\refstepcounter{figure}\label{fig:external}
\noindent{\large\bfseries Fig. \arabic{figure}\quad Clinical performance across validation settings, cancer types and horizons}\par
\begin{center}\includegraphics[width=\linewidth,height=0.73\textheight,keepaspectratio]{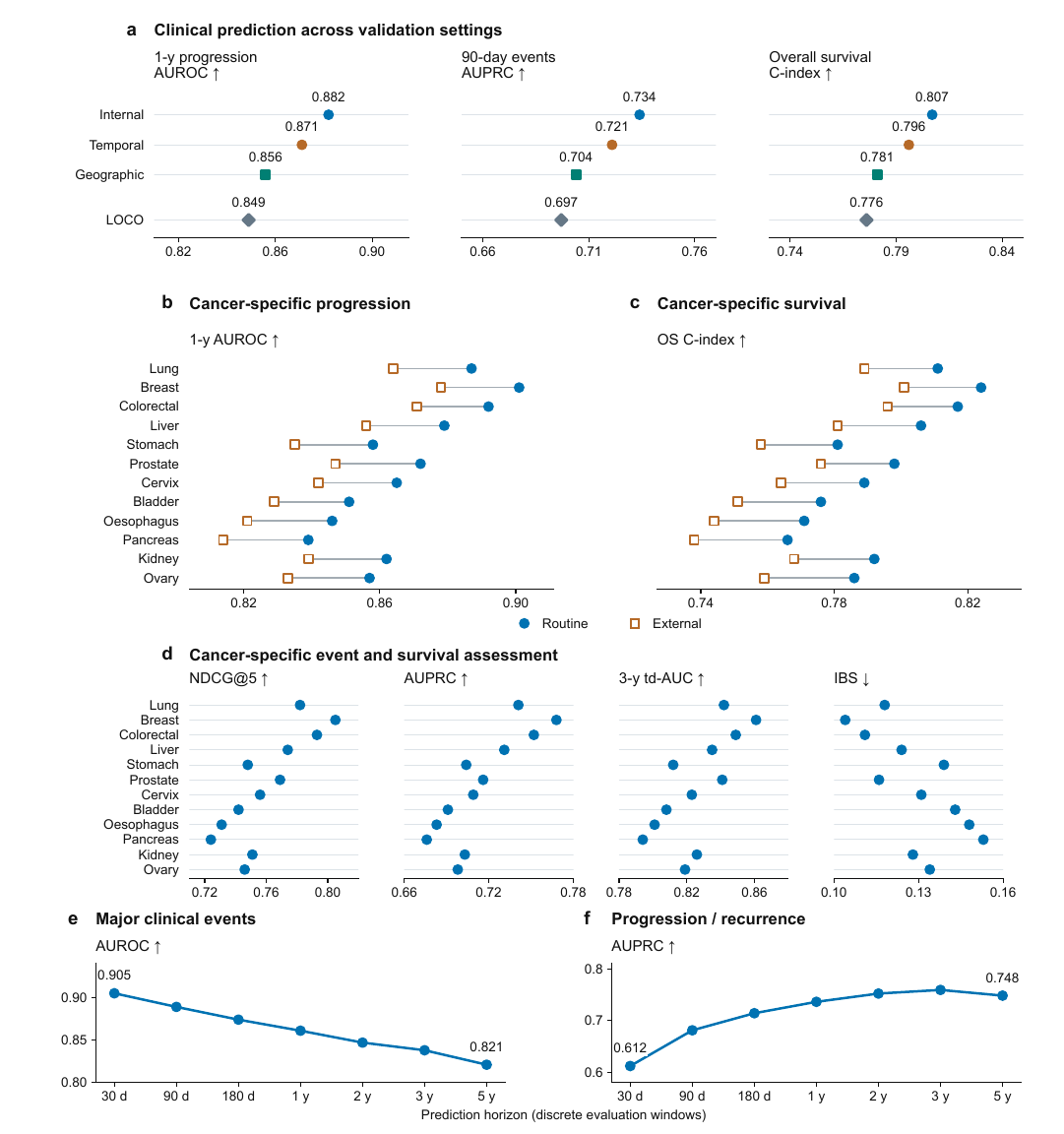}\end{center}
{\fontsize{8.5}{10.4}\selectfont a, One-year-progression AUROC, 90-day-event AUPRC and overall-survival C-index in internal, temporal-external, geographic-external and leave-one-centre-out (LOCO) evaluations. Circles denote internal and temporal-external results, squares denote geographic-external results, and diamonds denote LOCO means. b,c, One-year-progression AUROC and overall-survival C-index across twelve cancer types in routine and external evaluations. Blue filled circles and orange open squares denote routine and external evaluations, respectively; grey lines connect the two estimates for each cancer type. d, Next-event NDCG@5, 90-day-event AUPRC, three-year overall-survival time-dependent AUC and integrated Brier score (IBS), shown in the same cancer-type order; each column retains its own original scale. e,f, Next-major-clinical-event AUROC and progression/recurrence AUPRC at discrete evaluation horizons from 30 days to 5 years. Points show aggregate results at each horizon, and line segments connect adjacent horizons. Each row in b--d corresponds to one cancer type. Metrics are shown as aggregate point estimates. $\uparrow$ denotes higher is better; $\downarrow$ denotes lower is better.\par}
\end{minipage}}
\end{figure}

External evaluation showed that clinical readouts retained discrimination in new periods and medical centres (Fig.~\ref{fig:external}a). One-year progression AUROC was 0.871 and 0.856 in temporal and geographical external evaluations, respectively; 90-day event AUPRC was 0.721 and 0.704, and OS C-index was 0.796 and 0.781. Relative to internal testing, the three geographical external metrics declined by 0.026, 0.030 and 0.026. Leave-one-centre-out progression AUROC was $0.849\pm0.021$ (mean and between-centre standard deviation), revealing both cross-centre discrimination and centre heterogeneity.

Cancer-stratified analyses further characterize the range of diseases over which the model can be evaluated. In the cancer-specific evaluation sets, one-year progression AUROC ranged from 0.839 to 0.901 and C-index from 0.766 to 0.824; the corresponding cancer-specific external aggregates ranged from 0.814 to 0.878 and 0.738 to 0.801 (Fig.~\ref{fig:external}b,c). Cancer-specific event metrics, time-dependent survival AUC and integrated Brier scores are shown in Fig.~\ref{fig:external}d; age, sex and stage subgroups are presented in Table~\ref{tab:subgroups}.

Predictions across fixed horizons characterize the temporal range of this representation. From 30 days to 5 years, major-event AUROC declined from 0.905 to 0.821 (Fig.~\ref{fig:external}e). Progression/recurrence AUPRC across horizons is shown in Fig.~\ref{fig:external}f. These fixed-horizon evaluations complement free rollout evaluated by recursive step.

\begin{table}[!htbp]
\centering
\begin{minipage}{\textwidth}
{\fontsize{9}{11.5}\selectfont\setlength{\tabcolsep}{3pt}\renewcommand{\arraystretch}{1.25}

\caption{Prediction performance by age, sex and stage.}\label{tab:subgroups}
\begin{tabular}{@{}>{\raggedright\arraybackslash}p{0.15\linewidth}>{\raggedright\arraybackslash}p{0.23\linewidth}>{\raggedright\arraybackslash}p{0.2\linewidth}>{\raggedright\arraybackslash}p{0.18\linewidth}>{\raggedright\arraybackslash}p{0.18\linewidth}@{}}
\toprule
\textbf{Dimension} & \textbf{Subgroup} & \textbf{1-year progression AUROC} & \textbf{OS C-index} & \textbf{$\Delta$AUROC}\\
\midrule

Age & $<50$ years & 0.879 & 0.803 & -0.003\\
Age & 50--64 years & 0.884 & 0.809 & +0.002\\
Age & 65--74 years & 0.881 & 0.806 & -0.001\\
Age & $\geq75$ years & 0.873 & 0.798 & -0.009\\
Sex & Female & 0.885 & 0.810 & +0.003\\
Sex & Male & 0.879 & 0.804 & -0.003\\
Stage & I--II & 0.876 & 0.801 & -0.006\\
Stage & III & 0.883 & 0.808 & +0.001\\
Stage & IV & 0.888 & 0.812 & +0.006\\

\bottomrule
\end{tabular}

\noindent Age, sex and stage are separate stratifications. $\Delta$AUROC is subgroup one-year-progression AUROC minus the overall AUROC.\par}
\end{minipage}
\end{table}

\FloatBarrier

\FloatBarrier
\section{Discussion and outlook}

Saga connects future clinical events, timing and disease morphology through an updatable patient state. Paired follow-up describes future-image prediction, conditioning perturbations test the contribution of patient information, free-rollout comparisons show relative event and timing advantages, and multitask and hospital-external evaluations characterize the range of clinical readouts. These five groups of results organize multimodal generation and continuing trajectory prediction within a common patient-modelling framework.

Future phenotypes provide observable readouts beyond risk ranking. Multimodal oncology studies have demonstrated the complementary value of different observations for outcome prediction\citep{Jee2024,Vanguri2022,Xiang2025}; Saga further uses patient histories to predict subsequent disease appearances. Swapping or removing patient conditions while holding historical images fixed reduces event and morphology performance, supporting a contribution of patient information to future readouts. Ultrasound and MRI characterize lesion-level change, whereas pathology patches characterize local tissue phenotypes. These distinct evaluation scales correspond to lesion measurements, feature similarity and tissue-phenotype classification. Output differences under alternative treatment conditions describe model responses; causal interpretation requires an appropriate identification design\citep{Hernan2016}.

Event feedback allows the patient representation to update along a generated history. Generative health-record models have demonstrated sampling of future trajectories from patient histories\citep{Kraljevic2024foresight,Ethos2024,Shmatko2025}, and DT-GPT extends generation to longitudinal clinical variables\citep{DTGPT2025}. Sequential-prediction research has shown that errors can accumulate when later predictions depend on a model's own outputs\citep{Ross2011Sequential,Bengio2015}. Saga appends generated events and times and re-encodes the history at every step, maintaining event and timing advantages through eight steps in matched-origin comparisons. The event-F1 difference from the Transformer increased from 0.030 to 0.119. Edit distance, time-warping distance and feature drift also increased with rollout depth, indicating accumulated errors in longer trajectories. The two-reader pilot adds a preliminary assessment of trajectory plausibility and image--text--time consistency.

Examination frequency, modality availability and follow-up completeness in routine care depend on disease status and access to care. Electronic health record research has shown that the timing and frequency of laboratory testing can themselves carry predictive information\citep{Agniel2018biases}. Changes in long-horizon imaging error and evaluability should therefore be interpreted alongside acquisition protocols and tissue-sampling scales. Previous chest-radiograph AI studies have shown that predictive performance can vary across hospitals and patient populations\citep{Zech2018variable,SeyyedKalantari2021underdiagnosis}. Saga's hospital-external results and stratification by cancer, age, sex and stage further describe variation across patient settings (Fig.~\ref{fig:external}; Table~\ref{tab:subgroups}). The current aggregate results support descriptive comparisons of prediction and conditional responses.

\subsection{Outlook}

Further external-cohort validation will cover different populations, healthcare settings and longitudinal imaging data. Existing individual predictions will also be used to refine patient-level interval estimates for core comparisons and probability calibration for specified endpoints and horizons. Blinded assessments with more cases and physicians will examine trajectory plausibility and the value of assisted interpretation. Joint evaluation of events and images within the same patients will further characterize consistency and uncertainty across multiple possible futures.

In drug development and trial design, complete patient histories can be explored for estimating baseline risks, subsequent event rates and follow-up trajectories to support eligibility assessment, risk stratification and prognostic enrichment, with these predictions tested in independent trial data\citep{Liu2021TrialPathfinder,FDA2019Enrichment}. Patient selection based on treatment effects requires randomized trials or an explicit causal-identification design to investigate which patients are more likely to benefit from a particular intervention.

Clinical applications can progress from silent prospective evaluation to risk alerts, planning of reassessment timing and follow-up intensity, and patient-trajectory displays in multidisciplinary care. Studies embedded in clinical workflows will evaluate the effects of model information on physician decisions and patient outcomes\citep{Manz2023long}. Longer-term directions include extending oncology patient-state modelling to multimorbidity, post-treatment rehabilitation and health management across the life course, combining continuously updated examinations and behavioural information to study individual health-state evolution and population service needs\citep{Laubenbacher2024DigitalTwins}. Evidence for each use will be established through external validation and prospective studies in the corresponding target population.

\FloatBarrier\par\medskip\Needspace{10\baselineskip}\begingroup\interlinepenalty=10000
\begingroup\interlinepenalty=10000
\makeatletter
\renewcommand{\@biblabel}[1]{#1.}
\@ifundefined{bibnumfmt}{}{\renewcommand{\bibnumfmt}[1]{#1.}}
\makeatother
\renewcommand{\refname}{References}

\endgroup

\endgroup
\end{document}